\documentclass[conference]{IEEEtran}
\IEEEoverridecommandlockouts

\usepackage{graphics} 
\usepackage{epsfig} 
\usepackage{amsmath} 
\usepackage{amssymb} 
\usepackage{amsfonts}
\usepackage{mathtools}
\usepackage{algorithm}
\usepackage{array}
\usepackage{textcomp}
\usepackage{stfloats}
\usepackage{url}
\usepackage{verbatim}
\usepackage{graphicx}
\usepackage{cite}
\usepackage[inkscapeformat=png]{svg}
\usepackage{subcaption}
\usepackage[font={scriptsize}]{caption}
\usepackage[section]{placeins}
\usepackage{float}
\usepackage{booktabs}
\usepackage{makecell}
\usepackage{blindtext}
\usepackage{comment}
\usepackage{tablefootnote}
\usepackage{footnotehyper}
\usepackage[colorlinks,linkcolor=blue]{hyperref}

\graphicspath{{./figures/}}
\newcommand{\reals}{\mathbb{R}}
\usepackage{tcolorbox}
\usepackage{svg}
\usepackage{algpseudocode}
\usepackage{hyperref}

\tcbset{
  promptstyle/.style={
    colback=gray!10,    
    colframe=gray!70,   
    fonttitle=\bfseries,
    boxrule=0.5pt,
    arc=2mm,            
    left=2mm,
    right=2mm,
    top=1mm,
    bottom=1mm,
    enhanced,
    breakable           
  }
}

\begin{document}

\title{Graph-Fused Vision-Language-Action for Policy Reasoning in Multi-Arm Robotic Manipulation
}
\author{Shunlei Li$^{1,*}$, Longsen Gao$^1$, Jiuwen Cao$^2$, Yingbai Hu$^2$ 
\thanks{Shunlei Li and Jiuwen Cao are with Machine Learning and I-health International Cooperation Base of Zhejiang Province, Artificial Intelligence Institute, Hangzhou Dianzi University, Zhejiang, 310018, China. (e-mail: {\tt\small shunlei.li@outlook.com} ; {\tt\small jwcao@hdu.edu.cn})}
\thanks{Longsen Gao is with the Electrical and Computer Engineering Department, The University of New Mexico, Albuquerque, NM 87131, USA. (e-mail:{\tt\small lgao1@unm.edu})}
\thanks{Yingbai Hu is with the School of Computation, Information and Technology, Technical University of Munich, 85748 Germany. (e-mail:{\tt\small yingbai.hu@tum.de})}
}

\maketitle

\begin{abstract}
Acquiring dexterous robotic skills from human video demonstrations remains a significant challenge, largely due to conventional reliance on low-level trajectory replication, which often fails to generalize across varying objects, spatial layouts, and manipulator configurations. To address this limitation, we introduce Graph-Fused Vision–Language–Action (GF-VLA), a unified framework that enables dual-arm robotic systems to perform task-level reasoning and execution directly from RGB-D human demonstrations. GF-VLA employs an information-theoretic approach to extract task-relevant cues, selectively highlighting critical hand–object and object–object interactions. These cues are structured into temporally ordered scene graphs, which are subsequently integrated with a language-conditioned transformer to produce hierarchical behavior trees and interpretable Cartesian motion primitives. To enhance efficiency in bimanual execution, we propose a cross-arm allocation strategy that autonomously determines gripper assignment without requiring explicit geometric modeling. We validate GF-VLA on four dual-arm block assembly benchmarks involving symbolic structure construction and spatial generalization. Empirical results demonstrate that the proposed representation achieves over 95\% graph accuracy and 93\% subtask segmentation, enabling the language–action planner to generate robust, interpretable task policies. When deployed on a dual-arm robot, these policies attain 94\% grasp reliability, 89\% placement accuracy, and 90\% overall task success across stacking, letter-formation, and geometric reconfiguration tasks, evidencing strong generalization and robustness under diverse spatial and semantic variations.
\end{abstract}

\begin{IEEEkeywords}
Multi-arm manipulation, Vision-Language-Action models, scene graph, learning from demonstration
\end{IEEEkeywords}

\section{Introduction}



Developing robotic systems capable of robust, precise manipulation in unstructured, dynamic environments remains a core challenge for embodied AI: agents must perceive complex scenes, reason about physical interactions, and execute multi-step behaviors under uncertainty. Traditional vision-based control (e.g., visual servoing) relies on fixed sensor geometries and low-level visual cues, making it fragile to noise, occlusion, and contact; moreover, limited use of complementary tactile and proprioceptive modalities constrains generalization to real-world conditions \cite{cao2021six, wang2024hypermotion,liu2025hybridvla}. Recent Vision-Language-Action (VLA) models leverage large-scale multimodal pretraining to map language instructions to actions, enabling semantic reasoning and task-level flexibility beyond trajectory imitation \cite{brohan2023rt2, kim2024openvla, black2024pi0}. Yet current VLAs still struggle with precise manipulation and dynamic physical interactions, lacking structured inductive biases to capture fine-grained spatial–temporal relationships; as a result, they can produce physically inconsistent plans, especially in contact-rich or ambiguous scenarios \cite{shridhar2022cliport}.

Parallel approaches, such as reinforcement learning, imitation learning, and transfer learning, have shown potential but remain sample-inefficient, computationally demanding, or dependent on extensive domain-specific data~\cite{li2018deep, peng2020learning, jaquier2025transfer}. Information-theoretic methods offer principled tools for reasoning about uncertainty and relevance~\cite{goyal2019infobot}, yet their integration with high-level task planning has been limited. Consequently, no existing paradigm fully unifies structured physical interaction modeling with semantic task reasoning, which is essential for flexible and trustworthy robotic manipulation.

\begin{figure*}[!t]
    \centering
    \includegraphics[width=0.95\linewidth]{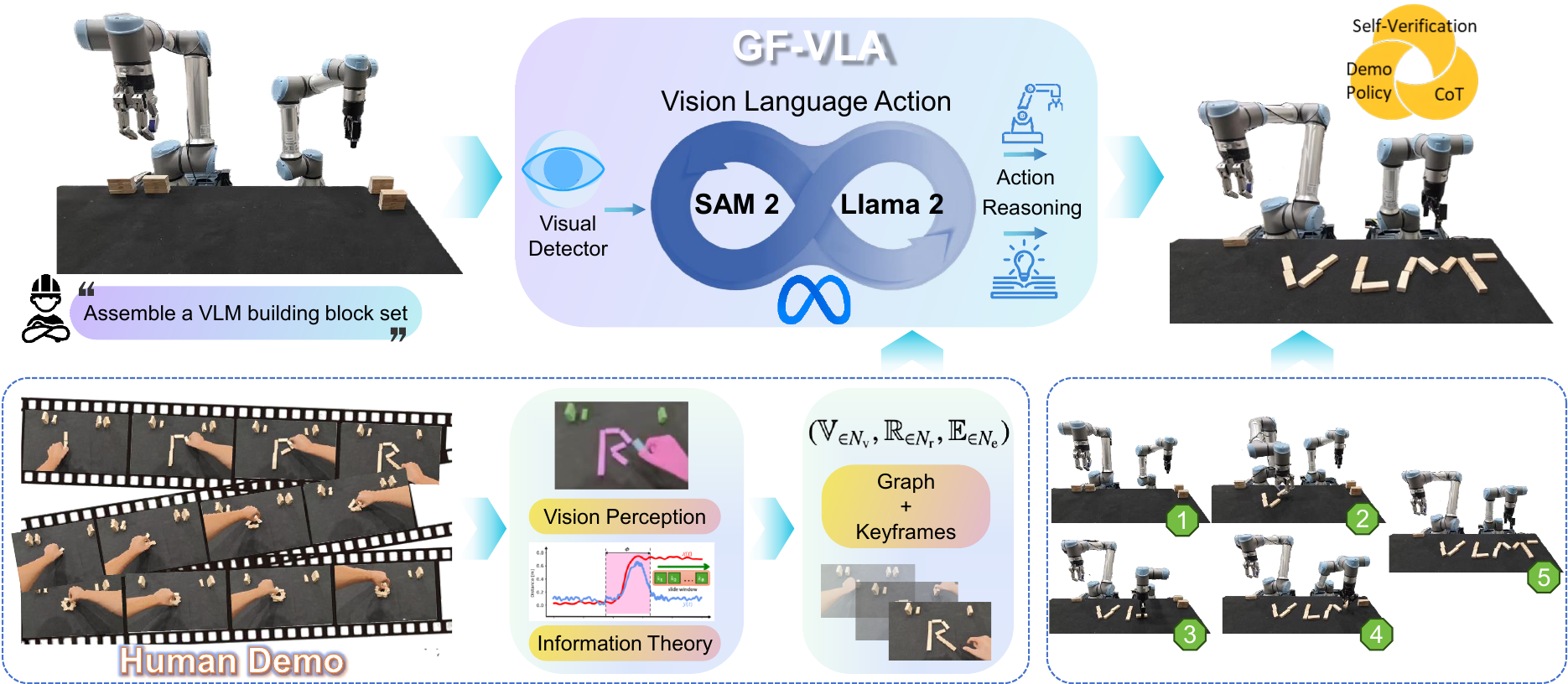}
    \caption{An overview of the GF-VLA framework that performs policy transfer from a single human demonstration to a dual-arm robot manipulation task. \textcolor{blue}{More detail: }\href{http://tiny.cc/iros25_vla}{http://tiny.cc/iros25\_vla}}
    \label{fig:overview_gf_vla}
\end{figure*}

To bridge this gap, we propose Graph-Fused VLA (GF-VLA), a framework that integrates structured scene graph representations with VLA reasoning as shown in Fig.~\ref{fig:overview_gf_vla}. Our method constructs temporally ordered, interaction-aware graphs from multimodal demonstrations using information-theoretic cues, and fuses them with a language-conditioned planner for semantically grounded task policies. Furthermore, we incorporate Chain-of-Thought prompting and self-verification to provide explicit subgoal decomposition and interpretable execution.

Our contributions are threefold:
\begin{itemize}
    \item We introduce an information-theoretic approach for constructing structured scene graphs from multimodal human demonstrations, explicitly encoding dynamic physical interactions.
    \item We propose GF-VLA, the first unified framework integrating structured interaction modeling with VLA reasoning, enabling robust and generalizable manipulation.
    \item We enhance interpretability by embedding Chain-of-Thought prompting into VLA models, providing transparent subgoal decomposition and improving execution reliability.
\end{itemize}

\section{Information-Theoretic Scene Graphs}
\label{sec::scene_rep}

\subsection{Information Theory in Robotic Manipulation}

Effective task representation hinges on isolating the scene’s active portion—regions exhibiting significant dynamical variation—thereby yielding compact, task-relevant abstractions while excluding inert tabletop elements. Information theory~\cite{shannon1948mathematical} offers a principled basis by quantifying signal information content independently of semantics; we measure activity via entropy, which captures the average uncertainty (unpredictability) of a signal. For a random variable $\mathcal{X}$ over the discrete support $\mathbb{X}_\phi$ within a temporal window $\phi$ with mass function $p(x)$ as

\begin{equation}
\label{eq:entropy}
    \mathcal{H}^\mathcal{X}(p) = -\epsilon \sum_{i=1}^{N_\mathrm{x}} p(x_i) \cdot \ln {p(x_i)},\ \epsilon \in \mathbb{R}_{+}
\end{equation}
where $\mathbb{X}_{\phi \in N_\mathrm{x}}{:=} \{x_1, \dots, x_{N_\mathrm{x}}\} = \{x{\scriptstyle (t_{\phi/2}-\tfrac{\phi}{2})}, \dots, x{\scriptstyle (t_{\phi/2}+\tfrac{\phi}{2})}\}$ in which $\phi$ denotes a sliding temporal window that we take to calculate entropy value $\mathcal{H}^{\mathcal{X}(t)}$ and $t$ denotes the time step within the temporal sliding window $\phi$; $x_i \in \mathbb{X}_{\in N_\mathrm{x}}$ is a discrete element from set $\mathbb{X}_{\in N_\mathrm{x}}$, and $\mathcal{X}:= x_i \sim \mathbb{X}_{\phi \in N_\mathrm{x}}$ denotes a random variable extracted from the set $\mathbb{X}_{\phi \in N_\mathrm{x}}$; $p(x_i)$ denotes the probability at $i_{\mathrm{th}}$ element in $\mathbb{X}_{\in N_\mathrm{x}}$, and $\epsilon \in (0,1)$ is a constant value. $\mathcal{H}^{\mathcal{X}(t_{\phi/2})}$ is the resulting entropy value, which is centered at the sliding window $\phi$. Note that Shannon's entropy is measured in bits. The higher the entropy value, the more bits are required to transmit the information contained in the signal. 

Within the proposed framework, entropy analysis is primarily applied to positional signals to characterize scene dynamics effectively. Rather than computing entropy for the entire signal simultaneously, the approach involves employing a sliding temporal window $\phi$. Specifically, entropy is calculated incrementally as this window traverses the signal over time. At each incremental shift of the temporal window, we use~\eqref{eq:entropy} to generate a temporal sequence of entropy $\mathcal{H}^{\mathcal{X}(t)}$. 

By evaluating patterns within this entropy-derived time series, it becomes possible to discern critical fluctuations in scene dynamics, thus identifying how and when significant changes occur throughout the task execution.

\begin{figure}[!t]
    \centering
    \captionsetup{font=footnotesize}
    \begin{subfigure}[c]{0.48\textwidth}
        \centering
        \captionsetup{font=footnotesize,,margin={0.6cm,0cm}}\includegraphics[width=0.99\linewidth]{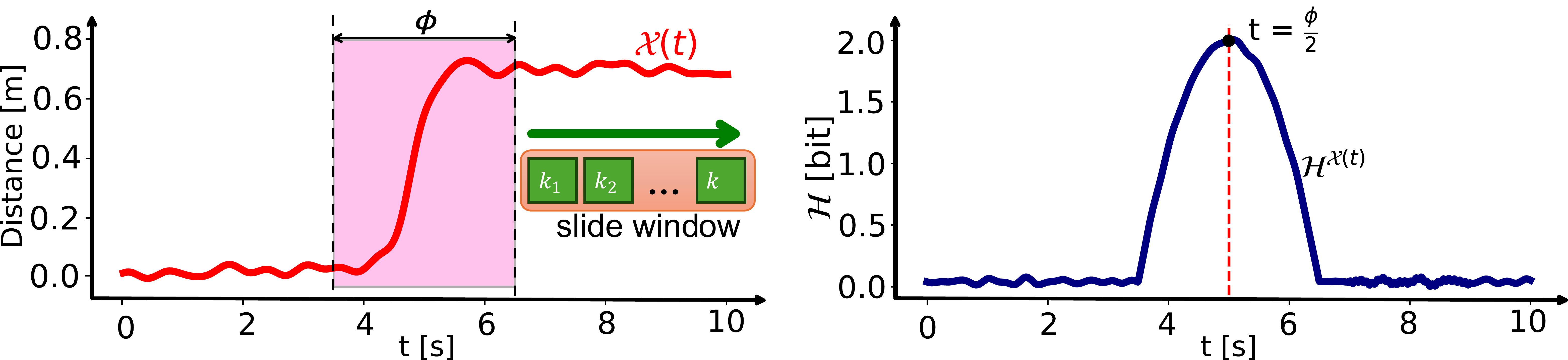}
        \subcaption{}
        \label{fig::entropy_1} 
        \includegraphics[width=0.99\linewidth]{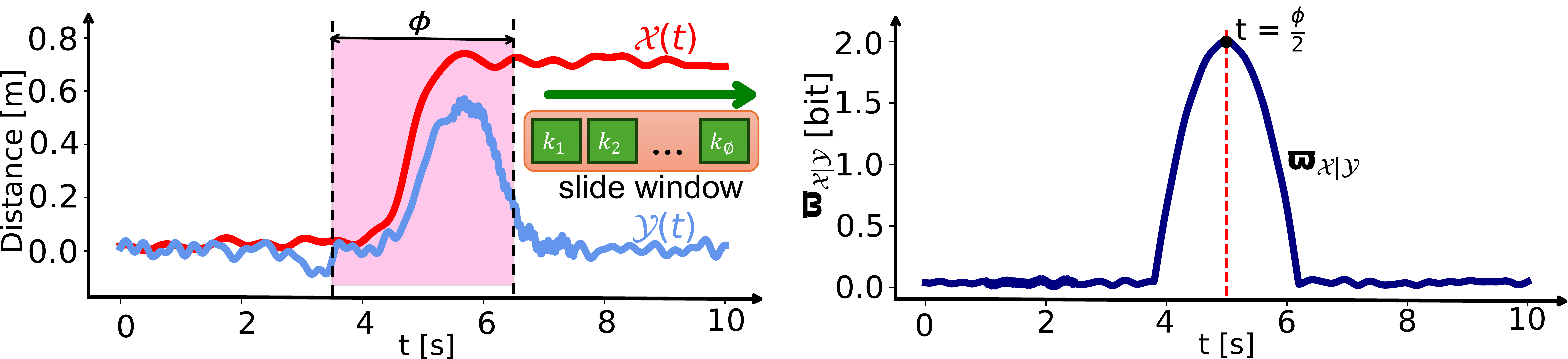}
        \subcaption{}
        \label{fig::entropy_2} 
    \end{subfigure}
    \caption{
     The relocation of a single object being manipulated over time. (a). The trajectory of a single object is shown, with the sliding window $\phi$ applied to the signal $\mathcal{X}(t)$ representing the object's position over time. (b) The entropy $\mathcal{H}^{\mathcal{X}(t)}$ is computed by sliding the window across $\mathcal{X}(t)$ and evaluating the entropy of the distribution of positions within each windowed segment. A bell-shaped curve emerges, highlighting periods of significant positional change.
    }
    \label{fig::entropy_all}
\end{figure}

We estimate the positional distribution $p(x_i)$ in each temporal window $\phi$ with a histogram: positions are quantized into fixed-width intervals $\zeta$, bin heights count occurrences, and—because $\zeta$ is fixed—the number of bins adapts to the observed positional range. Computing the associated entropy $\mathcal{H}(t)$ over $\phi$ exposes motion changes: uniform-speed motion yields a stable entropy profile, while deceleration increases predictability and lowers entropy. The quantization level $\zeta$ differentiates localized from extensive movement—constrained trajectories produce few high-probability bins (low entropy), whereas broad excursions populate many bins, approaching a near-uniform distribution (high entropy). Selecting $\phi$ large enough to span repeated cycles aggregates many identical samples into the same bin (sustained low entropy), but transient, nonrepetitive segments widen the distribution and create entropy peaks; e.g., stirring a liquid maintains low entropy, whereas subsequently placing the spoon several centimeters away induces a pronounced, short-lived rise. This purely statistical approach avoids noisy velocity/acceleration estimates from numerical differentiation. As in Fig.~\ref{fig::entropy_1}, a 1D position trace $\mathcal{X}(t)$ evaluated with a sliding window of width $\phi$ yields a typically unimodal, bell-shaped entropy profile $\mathcal{H}^{\mathcal{X}(t)}$; its peak at $t_{\phi/2}$ marks the event’s temporal center, while amplitude and duration reflect movement magnitude and speed. The derivative $\partial \mathcal{H}^{\mathcal{X}(t)}/\partial t$ further delineates dynamics: $\partial \mathcal{H}^{\mathcal{X}(t)}/\partial t>0$ signals initiation, and $\partial \mathcal{H}^{\mathcal{X}(t)}/\partial t<0$ indicates convergence to a new equilibrium.

A comprehensive analysis of manual tasks requires shifting from isolated scene elements to their interactions—reciprocal influences that imply information exchange. Shannon information theory offers a principled lens that we quantify shared information via mutual information $\boldsymbol{\varpi}$, which measures the statistical dependence between random variables $\mathcal{X}$ and $\mathcal{Y}$. Conceptually, $\boldsymbol{\varpi}$ captures the overlap of their uncertainties between the intersection of $\mathcal{H}^{\mathcal{X}}$ and $\mathcal{H}^{\mathcal{Y}}$ as

\begin{equation}
\boldsymbol{\varpi}_{\mathcal{X}|\mathcal{Y}} = \mathcal{H}^{\mathcal{X}} + \mathcal{H}^{\mathcal{Y}} - \mathcal{H}^{\mathcal{X},\mathcal{Y}}, 
\label{eq:mi}
\end{equation}
where $\boldsymbol{\varpi}_{\mathcal{X}|\mathcal{Y}}$ is called \emph{joint entropy} and represents the union of the two sets:
\begin{equation}
\label{eq:joint_e}
\boldsymbol{\varpi}_{\mathcal{X}|\mathcal{Y}} = - \sum_{i=1}^{N_\mathrm{x}} \sum_{j=1}^{N_\mathrm{y}} p(x_i,y_j) \cdot \ln{p(x_i, y_j)},
\end{equation}
where $x_i$ and $y_j$ are discrete measurements from $\mathbb{X}_{\phi \in N_\mathrm{x}}$ and $\mathbb{Y}_{\phi \in N_\mathrm{y}}$, respectively, and $p(x_i,y_j)$ is the joint probability of both events occurring simultaneously in the time window $\phi$. Based on \eqref{eq:mi}, we can conclude that if $\boldsymbol{\varpi}_{\mathcal{X}|\mathcal{Y}} = 0$, $\mathcal{X}$ is independent of $\mathcal{Y}$ and vice versa, due to the symmetry of this measure. In other words, knowledge about the value of one variable provides no information about the other variable. Since we are interested in observing the dynamics of $\boldsymbol{\varpi}$ over time, we apply the same approach described earlier, shifting the time window $\phi$ along the considered signals to obtain $\boldsymbol{\varpi}_{\mathcal{X}(t)|\mathcal{Y}(t)}$.

Applying mutual information $\boldsymbol{\varpi}$ to position signals quantifies kinematic dependence between entities, enabling robust detection of hand–object coupling in manual tasks. Unlike velocity-based correlations, $\boldsymbol{\varpi}$ captures both linear and nonlinear dependencies; for example, during transport of multiple items on a tray, it correctly treats the hand and payload as a single kinematic unit despite minor sliding or oscillations. In a canonical grasp-and-move sequence as shown in Fig.~\ref{fig::entropy_2}, the time-varying $\boldsymbol{\varpi}_{\mathcal{X}|\mathcal{Y}}$ between object $\mathcal{X}(t)$ and hand $\mathcal{Y}(t)$, computed via a sliding window, exhibits a distinct unimodal peak confined to the coordinated-movement interval and remains near zero during pre-grasp approach and post-release retreat—disambiguating genuine interaction from incidental proximity in cluttered scenes. For 3D analysis, we aggregate across axes: for hand $h\in\mathbb{H}$ and object $o_i\in\mathbb{O}$ (with cardinalities $N_h, N_o$), the total mutual information is
\begin{equation}
\boldsymbol{\varpi}_{h,o_i}=\sum_{a\in\{x,y,z\}}\boldsymbol{\varpi}_{h,o_i}^{(a)}.
\end{equation}

\begin{figure}
    \centering
    \includegraphics[width=1\linewidth]{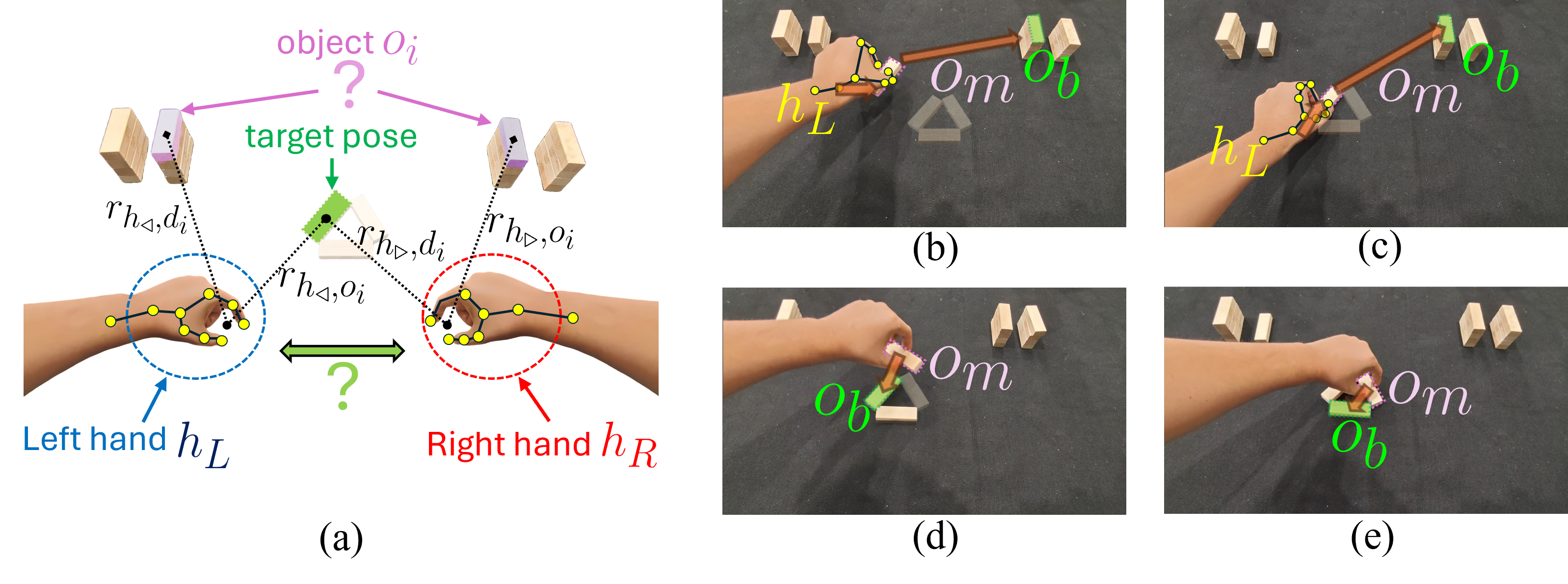}
    \caption{(a) The conceptual representation of the dual-hand selection policy. The framework depends on the priority of the Left hand $h_L$ or the Right hand $h_R$, which is optimal for interacting with the manipulated object $o_m$ to move it to the target pose. (b) denotes \texttt{Coupled-Motion} integration between only the left-hand $h_L$ and one manipulated Jenga block $o_m$. (c) denotes the \texttt{Docked} interaction between $h_L$ and one Jenga block $o_m$. (d) denotes \texttt{E-OO} interaction between the manipulated jenga block $o_{m}$ and one background jenga block $o_b$ on the table. (e) denotes the \texttt{T-OO} interaction between one manipulated jenga block $o_{m}$ and current background jenga block $o_b$ when the hand is shaken to make the building blocks shift slightly near the target position.}
    \label{fig:topo_4_relationship}
\end{figure}

\subsection{Scene Graph Generation}
We employ a scene graph, which is a directed graph data structure, to represent the system state and the relationships between its components. Formally, a graph can be denoted as $\texttt{SR}:= (\mathcal{G}, \mathcal{F})$ where $\mathcal{G}:=(\mathbb{V}_{\in N_\mathrm{v}}, \mathbb{R}_{\in N_\mathrm{r}}, \mathbb{E}_{\in N_\mathrm{e}})$ is a tuple in which $\mathbb{V}_{\in N_\mathrm{v}}$ is a set with $N_v$ number nodes, $\mathbb{R}$ is the set of relationships between the nodes with number $N_\mathrm{r}$, and $\mathbb{E}$ is a set of edges with number $N_\mathrm{e}$, and $\mathcal{F}:= f \sim \mathbb{K}_{\phi \in N_x}$ is a combined set of frames which we extract from the learning video during a temperoal sliding window. In our framework, the nodes $\nu_i:=(\lambda_i, \Lambda_i) \in \mathbb{V}_{\in N_\mathrm{v}}$ in which  $\Lambda_i$ is a set of attributes and $\lambda_i$ is its class identity, correspond to the entities within the scene, specifically hands and objects.  For our application, the sole attribute is the entity's 6D pose $\mathbf{q}_i \in \reals^6$. The directed edges $e_i \in \mathbb{E}_{\in N_\mathrm{e}}$ denote the interactions between these entities. Within the context of uni-manual tasks, interactions are classified into two principal categories:
\begin{itemize}
    \item \textit{\textbf{H}and-\textbf{O}bject (HO) Interactions}: denotes the relationship that pertains to the coupling between a hand and a manipulated object.
    \item \textit{\textbf{O}bject-\textbf{O}bject (OO) Interactions}: denotes the relationship between one moving manipulated object in hand and another static object in the background.
\end{itemize}

Interaction detection between any two entities instantiates a directed edge between their corresponding nodes in the scene graph. Generation proceeds by first identifying \textit{hand–object (HO)} interactions, reflecting the hand’s role as the primary effector of state change. \textit{Object–object (OO)} interactions are then evaluated only conditionally—i.e., if at least one \textit{HO} interaction was detected; otherwise, OO analysis is bypassed.

\subsubsection{Hand–Object Interactions Detection} We define two classes of Hand-Object ($\textit{HO}$) interactions as following:
\begin{itemize}
    \item \texttt{Coupled-Motion}: denotes interaction by the active displacement of an object by the hand, indicating that the hand holds the object and moves simultaneously.
    \item \texttt{Docked}: denotes a state of relative static contact between the hand and object, wherein the hand is in physical contact with an object but no joint movement is observed.
\end{itemize}

If the proximity condition is satisfied, the interaction is classified based on the mutual information, $\boldsymbol{\varpi}_{h,o_i}$. An active \texttt{Coupled-Motion} is declared if $\boldsymbol{\varpi}_{ho_i}$ is greater than a near-zero threshold $\alpha_{\boldsymbol{\varpi}}$, signifying strong kinematic coupling. Conversely, a \texttt{Docked} interaction is established or maintained if either the mutual information is concurrently decreasing and below the threshold (i.e., $\frac{\partial \boldsymbol{\varpi}_{h,o_i}}{\partial t} < 0$ and $\boldsymbol{\varpi}_{h,o_i} < \gamma{\boldsymbol{\varpi}}$), or if a \texttt{Docked} state was already present in the previous frame $k-1$. Any established interaction is terminated once the distance $\bar{r}_{h,o_i}$ subsequently exceeds the threshold $r^{\mathrm{th}}_{h,o} \in \reals_{\ge 0}$.


By design, the state $\texttt{Docked}$ is entered only following a $\texttt{Coupled\text{-}Motion}$ phase, preventing transient, non-manipulative contacts from being misclassified as salient interactions and thereby improving robustness in cluttered scenes. Detection then proceeds iteratively over all objects $o_k\in\mathbb{O}$ ($|\mathbb{O}|=N_o$), ordered by increasing proximity to the hand $h$; the search for hand–object coupling halts at the first positive detection, after which the pipeline advances to object–object (OO) interaction analysis.

\subsubsection{Object–Object Interactions Detection}
We define two classes of Object-Object ($OO$) interactions in static as following:
\begin{itemize}
    \item \texttt{Efficient OO (E-OO)}: denotes an interaction between a manipulated object $o_m$ and a stationary background object $o_b$ that is identified as being stable, intentional, and integral to the execution of a task. This type of interaction is considered essential and is subsequently mapped into robot command primitives during plan generation.
    \item \texttt{Transitory OO (T-OO)}: denotes incidental interaction between a manipulated object $o_m$ and a stationary background object $o_c$ that is identified as being unstable, transitory, and not representative of the core activity. These interactions are considered non-essential and are filtered out during the task segmentation process, meaning they are not translated into robot commands.
\end{itemize}

Static $OO$ interactions require spatial proximity: the mean distance $\bar r_{o_m,o_b}$ must satisfy $\bar r_{o_m,o_b}<r^{\mathrm{th}}_{\mathrm{oo}}$. When this holds, an \texttt{E-OO} interaction is established or confirmed if (i) a \texttt{Docked} relation exists between the hand $h$ and manipulated object $o_m$, or (ii) \texttt{E-OO} between $o_m$ and $o_b$ was present at frame $k-1$; thus, post-manipulation holding of $o_m$ near $o_b$ is correctly treated as a salient static relation. During \texttt{Coupled-Motion} (active manipulation), mutual information is ill-suited when a stationary object has near-zero positional entropy, so we instead use the entropy of the mean distance, $\mathcal{H}(\bar r_{o_m,o_b})$: the $OO$ interaction is labeled \emph{efficient} if $\partial \mathcal{H}(\bar r_{o_m,o_b})/\partial t<0$ (converging to a stable configuration) and \emph{Transitory} otherwise; the \texttt{E-OO} label persists unless $\bar r_{o_m,o_b}$ later exceeds $r^{\mathrm{th}}_{\mathrm{oo}}$. For example, moving $o_i$ past $o_{i-1}$ produces a rise in distance entropy after initial proximity, yielding a \emph{Transitory} label before termination by the distance threshold. Evaluation iterates over candidate background objects $o_b$, ordered by increasing distance from $o_m$, and stops at the first detected $OO$ relation; requiring all static $OO$ edges to involve $o_m$ prevents spurious links among incidental tabletop objects and stabilizes graph topology in clutter. After integration, $\mathcal{SR}[k]$ matches one of the topologies in Fig.~\ref{fig:topo_4_relationship}. If both a hand–object ($HO$) and a subsequent \texttt{RS-OO} relation are detected, the graph at time $t$ is fully specified by $\mathbb{V}_t=\{h,o_1,o_2\}$ with $h=(\mathcal{L}_h,\mathbf{p}_h(t))$, $o_1=(\mathcal{L}_{o_1},\mathbf{p}_{o_1}(t))$, $o_2=(\mathcal{L}_{o_2},\mathbf{p}_{o_2}(t))$; directed edges $\mathbb{E}_t=\{e_{h\rightarrow o_1},e_{o_1\rightarrow o_2}\}$; and relations $\mathbb{R}_t=\{r_{h\rightarrow o_1},r_{o_1\rightarrow o_2}\}$, where $r_{h\rightarrow o_1}$ is annotated with its type and $\varpi_{h,o_1}(t)$, and $r_{o_1\rightarrow o_2}$ with its static type. If no static $OO$ relation is found, the graph reduces to the $HO$ pair: $\mathbb{V}=\{h,o_1\}$, $\mathbb{E}=\{e_{h\rightarrow o_1}\}$, $\mathbb{R}=\{r_{h\rightarrow o_1}\}$.

\subsubsection{Dynamic Learning Policy of Dual-Hand Selection}

Dual‑arm execution requires allocating the grasping role to one of two hands so as to (i) reduce cumulative reach–grasp–place cost and (ii) mitigate inter‑arm interference. We formulate a dynamic, learning‑augmented selector that is kinematically grounded by a contralateral prior and refined from demonstrations. For the $i$‑th placement subtask with target pose $o_i$ (near the workspace midline) and source stack $d_i$ (left/right piles), we define the hand‑selection state at time $t$ as
\begin{equation}
\begin{aligned}
s_t &\triangleq 
\big\{ r_{h_L,d_i},\, r_{h_R,d_i},\, r_{h_L,o_i},\, r_{h_R,o_i} \big\}\\
r_{h,p} &= \big\|p - p_h(t)\big\|_2,
\label{eq::hand‑selection_state}
\end{aligned}
\end{equation}
where $h_L, h_R$ denote the left/right end‑effectors with Cartesian positions $p_h(t)$. The action space is

\begin{equation}
    \mathcal{A}=\{\textsc{use‑left‑hand},\,\textsc{use‑right‑hand}\}.
    \label{eq::hand_choose}
\end{equation}

We adopt a deterministic prior policy that embodies the crossing‑arm strategy: grasp with the hand contralateral to the target placement side. Equivalently, choose the hand opposite the one closer to $o_i$ as
\begin{equation}
    \pi_0(s_t) =
\begin{cases}
\textsc{use‑left‑hand}, & r_{h_R,o_i} < r_{h_L,o_i},\\[2pt]
\textsc{use‑right‑hand}, & r_{h_L,o_i} \le r_{h_R,o_i}.
\end{cases}
\label{eq::hand_choose_simple}
\end{equation}

This rule shortens the combined reach–grasp–place path and lowers cross‑arm interference during placement. Let $a_t^\star\in\mathcal{A}$ denote the grasping hand used by the human demonstrator at decision point $t$ (extracted from the \textit{HO} graph), and let $\pi_\theta(a\mid s_t)$ be a stochastic selector as a two‑way MLP classifier that outputs a distribution over $\mathcal{A}$. We provide an imitation‑shaped scalar signal

\begin{equation}
    R(a_t,a_t^\star) \;=\;
\begin{cases}
+R_{\text{bonus}}, & a_t = a_t^\star,\\
-R_{\text{penalty}}, & a_t \neq a_t^\star,
\end{cases}
\quad R_{\text{bonus}},R_{\text{penalty}}>0,
\label{eq::imitation-shaped_signal}
\end{equation}

Note that the train $\pi_\theta$ by weighted cross‑entropy (or equivalently, bandit‑style policy gradient with \eqref{eq::imitation-shaped_signal} as advantage). Concretely, we minimize

\begin{equation}
    \mathcal{L}_t(\theta) =  
w_+\,\mathbf{1}[a_t^\star]\,-\,w_-\,\mathbf{1}[\bar a_t^\star]
\cdot \left(-\log \pi_\theta(a_t^\star\mid s_t)\right),
\label{eq::minimize_eq}
\end{equation}
where $w_+=R_{\text{bonus}}$ and $w_-=R_{\text{penalty}}$, and $\bar a_t^\star$ the complement action. This yields a selector aligned with expert preferences while remaining robust to geometric variations. At run time, we fuse the contralateral prior \eqref{eq::hand_choose_simple} with the learned selector by biasing the logits toward the prior action:

\begin{equation}
\begin{aligned}
J(a\mid s_t) &= \log \pi_\theta(a\mid s_t)\;+\;\kappa\,\mathbf{1}\!\left[a=\pi_0(s_t)\right],\\
a_t &= \arg\max_{a\in\mathcal{A}} J(a\mid s_t),
\end{aligned}
\label{eq::logits_eq}
\end{equation}
where $\kappa\ge 0$ controls trust in the prior (recovering the deterministic rule as $\kappa\to\infty$). When an \texttt{E‑OO} Docked relation is active (the manipulated object is stably held against its background support), the current hand assignment persists to avoid mid‑task role flips; otherwise (4) is re‑evaluated at each decision point.


\section{Graph-Fused VLA (GF-VLA)}


\subsection{System architecture}
\label{framework}

Fig.~\ref{fig:overview_gf_vla} presents a two-stage framework: (i) Policy Generation from Human Demonstrations and (ii) VLA-based Execution with Chain-of-Thought (CoT) Reasoning. In Stage 1, raw videos are transformed into information-theoretic behavior graphs and temporal keyframes that encode spatial–semantic structure; a Policy Agent reasons over these graph–frame abstractions to produce a robot-executable policy—an ordered subtask sequence with an explicit reasoning trace—under a CoT + self-verification scheme. In Stage 2, the demonstration-derived CoT policy, live RGB input, and spoken instructions are fused by our GF-VLA model to adapt the policy to the current scene. The model outputs (a) an action head for low-level commands and (b) an LLM head for hierarchical planning, subtask decomposition, and CoT-based reasoning. During execution, continuous self-verification tracks progress, detects inconsistencies, and triggers local replanning, yielding robust, interpretable, and adaptive multi-step manipulation in unstructured environments.


\begin{figure*}[!t]
    \centering
    \includegraphics[width=0.95\linewidth]{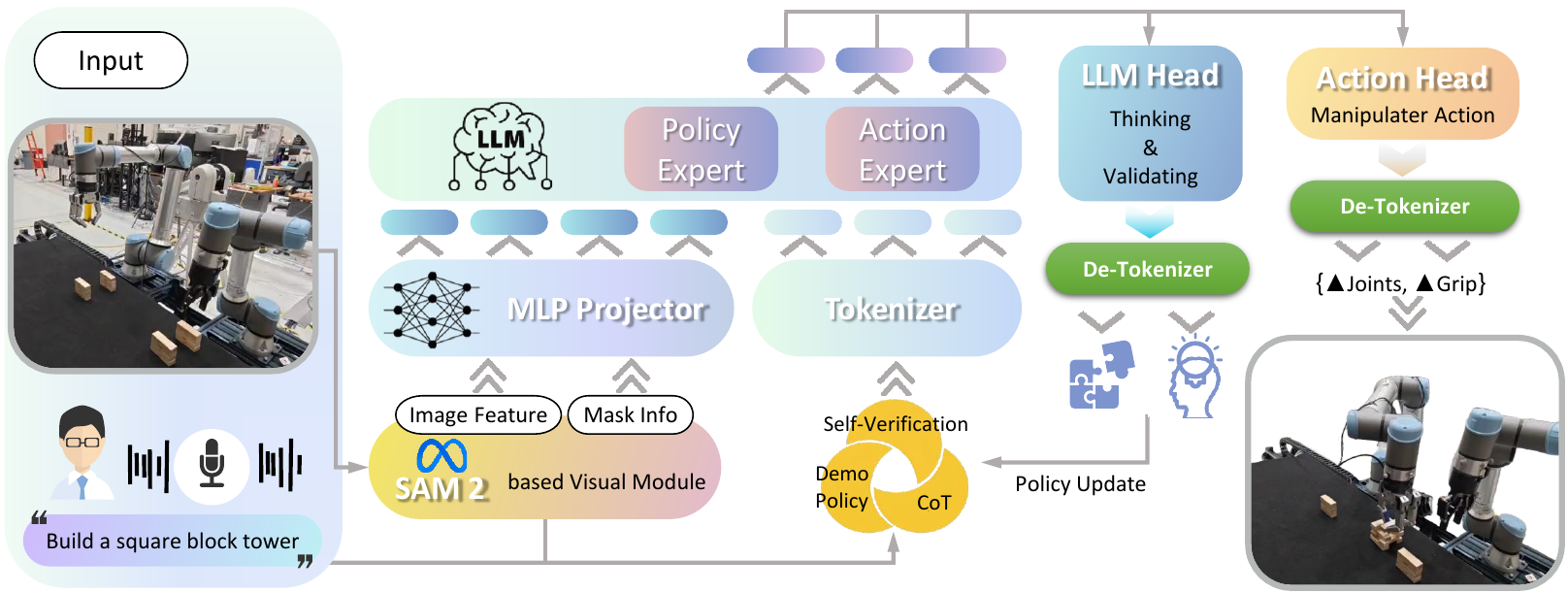}
    \caption{Policy transfer from a single human demonstration to a novel dual-arm robotic assembly task. The framework processes multimodal inputs, including language commands and visual scene data, using a SAM 2-based module to extract and project features into a shared embedding space. At its core, a unified Large Language Model (LLM) employs a dual-head structure: the LLM Head performs high-level semantic planning and validation using Chain-of-Thought (CoT) and self-verification, while the Action Head generates low-level, executable manipulator actions. This integrated design enables the robot to de-tokenize abstract reasoning into physically grounded joint and gripper commands, translating high-level goals into robust, real-world execution.}
    \label{fig:VLA_overview}
\end{figure*}

\subsection{Unified Dual-Head Architecture for Vision-Language-Action Reasoning}
\label{sec:model_architecture}
GF-VLA extends the standard VLA paradigm with four components: (i) a visual module that segments RGB and encodes RGB-D with binary masks into patch-level embeddings; (ii) a lightweight two-layer MLP projector for cross-modal token alignment; (iii) a language module built on a pretrained 7B-parameter LLaMA-2 backbone adapted with the Open X-Embodiment corpus \cite{kim2024openvla}, augmented for hierarchical policy generation, Chain-of-Thought (CoT) decomposition, and self-verification; and (iv) two task-specific output heads—an LLM Head for structured semantic planning and interpretable reasoning, and an Action Head for low-level motion control. Unlike prior VLA designs, GF-VLA integrates both heads within a unified transformer encoder, enabling parallel reasoning and action under a shared semantic representation. As shown in Fig.~\ref{framework}, top-down RGB-D frames and transcribed voice instructions (via a Gladia ASR module) are fused by projecting visual patches and masks through the MLP and concatenating them with prompt tokens to the language backbone. At run time, the LLM Head emits a one-shot structured reasoning trace, while the Action Head streams 5 Hz control commands—Cartesian end-effector poses and gripper states—decoded and dispatched to the corresponding robotic arms.

\subsection{Chain-of-Thought Guided Semantic Policy Planning}
\label{sec:policy_with_cot}

Robots acting in the open world must produce plans with logical structure, causal coherence, and commonsense alignment, yet LLM-based planners often default to “fast thinking”—single-step, reactive outputs ill-suited to multi-stage manipulation. We therefore integrate Chain-of-Thought (CoT) reasoning into GF-VLA’s policy generation \cite{wei2022chain}, prompting the Planning Agent to explicitly articulate intermediate reasoning, decompose high-level goals into interpretable subgoals, and slow decision-making for greater logical consistency, correctness, and auditability. Concretely, the CoT-enhanced agent yields a behavior tree aligned with human task logic whose nodes specify action type and parameters together with step-wise rationale and self-verification criteria. In constructing the letter “R,” for example, the robot first executes a dual-arm grasp (\texttt{PickObjDual}) to acquire two foundational blocks for the vertical backbone, then performs a coordinated dual-arm placement (\texttt{PlaceObjDual}) to form the diagonal leg; each node includes a reasoning statement and a verification clause grounded in sensor feedback (e.g., gripper status, object pose, scene stability). This interpretable, CoT-guided plan supports temporal consistency, symbolic understanding, and robust execution in complex bimanual manipulation tasks.

\subsection{Parameter-Efficient Multi-Head Fine-Tuning with LoRA}
\label{sec:training}

We adopt parameter-efficient fine-tuning via LoRA \cite{hu2021lora}, inserting separate adapters for the LLM and Action Heads within shared LLaMA-2 transformer blocks while freezing the visual encoder and projector; only the LLaMA backbone and LoRA adapters are updated. The LLM Head is trained with policy-level supervision from human demonstrations: 250 RGB videos by 10 participants performing letter-based symbolic assembly and tower construction \cite{merlo2025exploiting}. Videos are segmented into graph-based keyframes and paired with expert strategies; 125 demonstrations supervise next-token prediction over plan steps and Chain-of-Thought (CoT) reasoning, and 125 are held out for policy-generation evaluation (Subsec.\~5.2).

The Action Head is fine-tuned on 240 bimanual trials collected on a dual-arm platform—UR5e + Robotiq 2F-85 and UR10e + Barrett BH282—using a top-mounted Intel RealSense D435i for overhead RGB-D perception. Four task classes capture core skills: (i) shape generalization across cuboid/triangular-prism/cube; (ii) spatial relations under ambiguous instructions; (iii) absolute 6D pose execution; and (iv) relative pose execution. Each class × 3 shapes × 20 reps yields $4\times3\times20=240$ trials, recorded as synchronized dual-arm sequences with role-specific assignments and 6-DoF end-effector targets; supervision uses regression losses on Cartesian poses and gripper states. We split 120/120 for fine-tuning/evaluation. Training alternates policy and action batches, activating the corresponding LoRA branch and loss; GF-VLA fine-tunes in \~40 h on a single NVIDIA RTX 4090 and attains a 7.56 s average instruction-inference latency with bfloat16, enabling interpretable reasoning and robust dual-arm manipulation across diverse spatial configurations.


\section{Experimental Results}
\subsection{Task representation and planning}
\label{subsec:exp1_video}

\begin{figure}[!t]
  \centering
\captionsetup[sub]{font=small}
  \begin{subfigure}[t]{0.49\linewidth}
    \includegraphics[width=\linewidth]{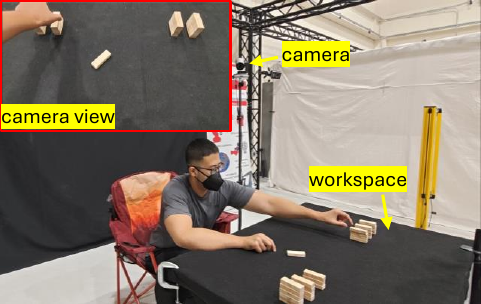}
    \caption{}
    \label{fig:left}
  \end{subfigure}\hfill
  \begin{subfigure}[t]{0.49\linewidth}
    \includegraphics[width=\linewidth]{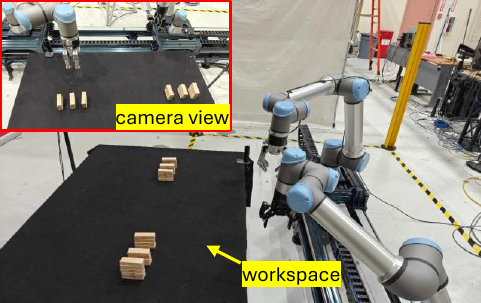}
    \caption{}
    \label{fig:right}
  \end{subfigure}
  \caption{(a) Configuration of the experimental environment and the associated camera viewpoint during the human demonstration phase. (b) The dual-arm robot system workspace and the visual feed are captured as the robot executes the task.}
  \label{fig:two}
\end{figure}

\begin{figure*}[!t]
    \centering
    \includegraphics[width=1\linewidth]{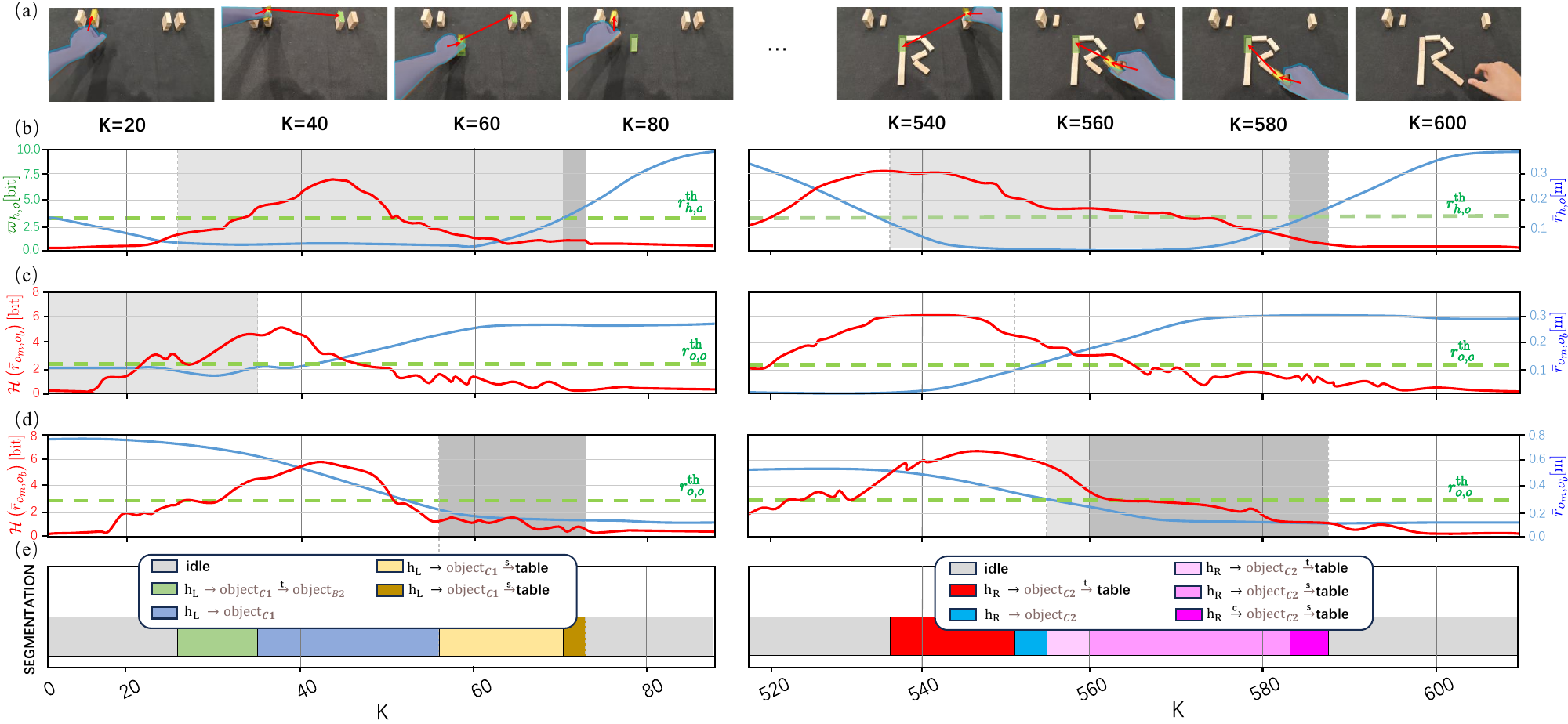}
    \caption{Information-theoretic analysis and temporal segmentation of a human demonstration for a block assembly task. (a) Keyframes from the demonstration video showing the human assembling the letter ``R". (b) Analysis of the Hand-Object (HO) interaction, plotting the mutual information ($\boldsymbol{\varpi}_{h,o}$) and relative distance ($\overline{r}_{h,o}$) between the hand and the manipulated block to detect kinematic coupling. (c)-(d) Analysis of Object-Object (OO) interactions, plotting the entropy of the relative distance ($\mathcal{H}(\overline{r}_{o_m,o_b})$) and the distance itself ($\overline{r}_{o_m,o_b}$) between the in-hand block and other objects to identify stable placements. (e) The resulting temporal segmentation of the task classifies the demonstration into a sequence of distinct interaction primitives based on the information-theoretic metrics.}
    \label{fig:demon_R_res}
\end{figure*}

The first experiment aimed to assess the functionality of the video processing module and the quality of the task representations derived from it. These representations served as input to the LLM, forming the foundation for downstream task planning and policy generation.

In the dual-hand construction of the letter “R”  as shown in Fig.~\ref{fig:demon_R_res}, the participant first chose a hand ($h_L$ or $h_R$) and placed $\mathrm{object_{C1}}$ drawn from two visible groups. During motion, $\mathrm{object_{C1}}$ briefly formed an $OO$ relation with another $\mathrm{object_{C1}}$ on the table at $k=40$ ($\bar r_{o,o}$ below threshold), but the rising distance-entropy $\mathcal H(\bar r_{o,o})$ labeled it \texttt{T-OO}, indicating that explicit table encoding was unnecessary. In alternate trials with slightly different initial $\mathrm{object_{C1}}$ poses, the hand skirted a box edge (treated as an obstacle), yielding a persistent \texttt{E-OO} that lasted until separation at $k=80$, coincident with a topology change. Approaching the table, a further \texttt{E-OO} between $\mathrm{object_{C1}}$ and the workspace was detected at $k=70$ via decreasing $\mathcal H(\bar r_{o,o})$ (descending red curve). No scene graph was emitted during the weighing phase because the hand disengaged from $\mathrm{object_{C1}}$. After re-grasping $\mathrm{object_{C2}}$, a brief proximity-induced \texttt{T-OO} with the workspace dissolved as $\mathcal H(\bar r_{o,o})$ rose; the scene-graph configuration at $K=552$ then remained stable until $\mathrm{object_{C2}}$ moved away. When $\mathrm{object_{C2}}$ encountered the previously placed block, \texttt{T-OO} transitioned to \texttt{E-OO} as $\mathcal H(\bar r_{o,o})$ decreased (dark gray region), with overall topology unchanged from $K=560$–$585$.

\begin{figure}[t]
  \centering
  \captionsetup[sub]{font=small}
  \begin{subfigure}[t]{0.50\linewidth}
    \includegraphics[width=\linewidth]{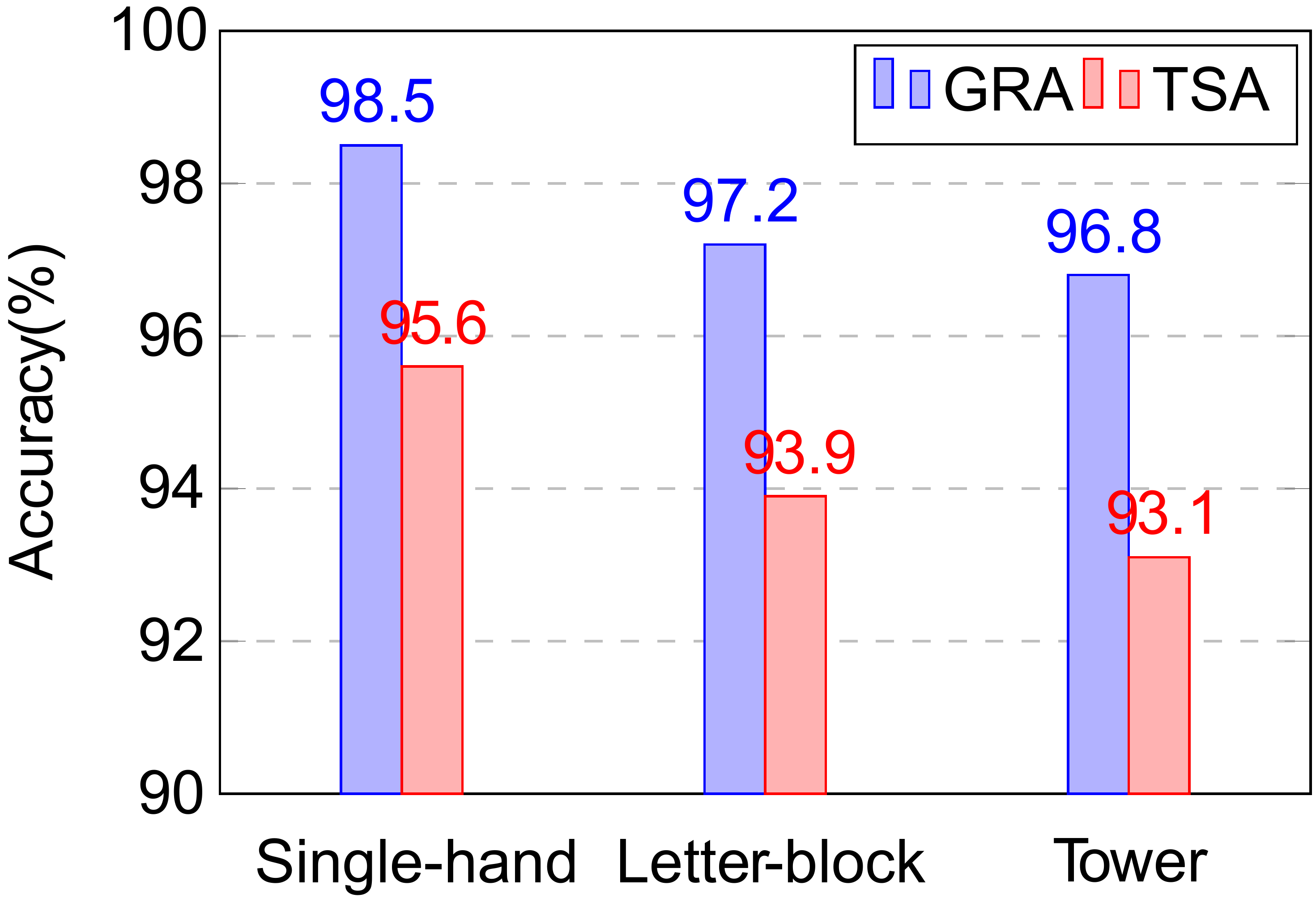}
    \captionsetup{margin={2em,0pt}} 
    \caption{}
    \label{fig:gra_tsa_bar}
  \end{subfigure}\hfill
  \begin{subfigure}[t]{0.44\linewidth}
    \includegraphics[width=\linewidth]{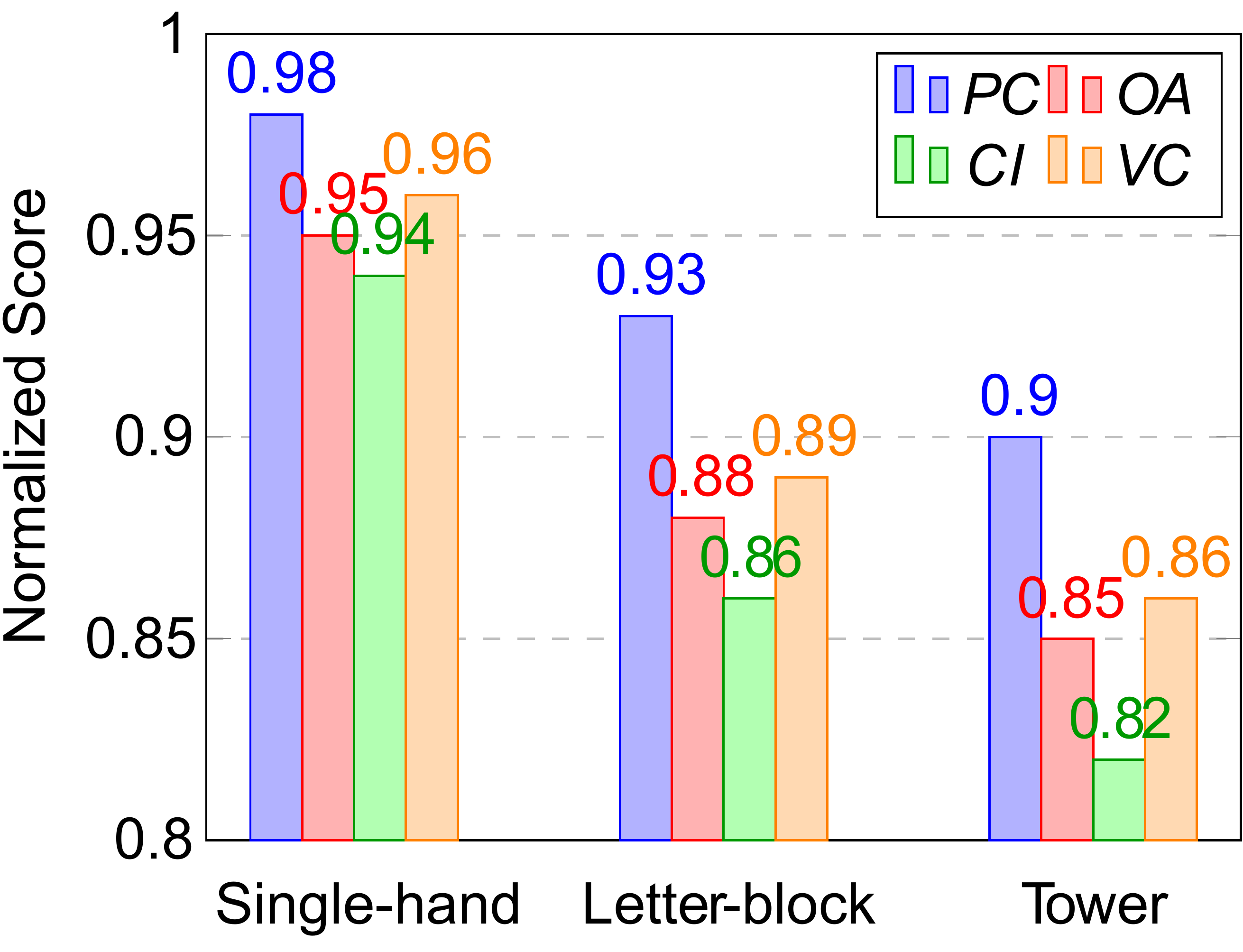}
    \captionsetup{margin={2em,0pt}} 
    \caption{}
    \label{fig:vla_llm_bar}
  \end{subfigure}
  \caption{(a) Comparison of Graph Representation Accuracy (GRA) and Task Segmentation Accuracy (TSA) across different task types. (b) LLM-generated plan and reasoning evaluation scores across tasks. All scores normalized to [0, 1] for visualization.}
  \label{fig::acc_plus_score}
\end{figure}

Fig.~\ref{fig:gra_tsa_bar} shows that the proposed method sustains high performance across task types for both Graph Representation Accuracy (GRA) and Task Segmentation Accuracy (TSA). In single-hand manipulation—characterized by low interaction complexity and well-separated actions—the system attains the highest scores (GRA 98.5\%, TSA 95.6\%). In letter-block assembly, despite bimanual coordination and dense layouts, GRA remains 97.2\% and TSA 93.9\%; errors are primarily due to temporal overlaps and ambiguous onsets of placement actions. Tower construction introduces vertical stacking, depth occlusion, and rapid execution; while GRA stays strong at 96.8\%, TSA drops to 93.1\%, indicating that temporal segmentation is more sensitive to rapid transitions and occlusions than spatial graph inference. Overall, Fig.~\ref{fig:gra_tsa_bar} indicates robust extraction of spatial and temporal structure with graceful degradation as task complexity increases.

Fig.~\ref{fig:vla_llm_bar} further shows that the VLA model’s LLM head produces accurate plans and coherent reasoning traces. For single-hand tasks, plan coverage is 98\%, ordering accuracy 95\%, and verification correctness 96\%, with chain-of-thought (CoT) interpretability rated 4.7/5. In letter-block assembly, coverage reaches 93\% and ordering 88\%; minor inversions in symmetric subtasks do not affect task validity, while CoT remains clear (4.3/5) and verification is 89\%. Tower construction yields 90\% coverage and 85\% ordering; CoT interpretability is 4.1/5 with occasional overgeneralization, and verification is 86\%. Collectively, these results demonstrate strong inference of subgoal structure, temporal dependencies, and causal rationales, while highlighting opportunities for improved fine-grained temporal modeling and object-level reasoning to further enhance generalization and robustness.

\subsection{Block manipulation assessment}

\begin{table}[t]
  \caption{Performance of Dual-Arm Manipulation under Different Task Conditions}
  \label{tab:gfvla_manipulation_results}
  \centering
  \setlength{\tabcolsep}{3pt} 
  \renewcommand{\arraystretch}{1.1}
  \footnotesize
  \resizebox{\columnwidth}{!}{%
  \begin{tabular}{@{}lcccc@{}}
    \toprule
    \textbf{Task Type} & \textbf{GSR} & \textbf{PSR} & \textbf{6DPE (cm, $^\circ$)} & \textbf{ICS} \\
    \midrule
    Shape Generalization         & 0.98 $\pm$ 0.02 & 0.95 $\pm$ 0.03 & 1.2 + 2.5 & 4.8 $\pm$ 0.3 \\
    Spatial Relation (Amb.)      & 0.94 $\pm$ 0.03 & 0.90 $\pm$ 0.04 & 2.1 + 4.2 & 4.5 $\pm$ 0.5 \\
    Absolute 6D Pose Exec.       & 0.92 $\pm$ 0.04 & 0.87 $\pm$ 0.05 & 2.8 + 6.0 & 4.2 $\pm$ 0.6 \\
    Relative Pose Exec.          & 0.91 $\pm$ 0.04 & 0.85 $\pm$ 0.06 & 3.1 + 6.8 & 4.4 $\pm$ 0.4 \\
    \midrule
    \textbf{Overall Average}     & \textbf{0.94}   & \textbf{0.89}   & \textbf{2.3 + 4.9} & \textbf{4.5} \\
    \bottomrule
  \end{tabular}
 } 
\end{table}

Table~\ref{tab:gfvla_manipulation_results} confirms that GF-VLA executes complex dual-arm manipulations with high success and spatial precision across diverse conditions. In shape generalization, grasp/placement success reached 98\%/95\% with minimal pose error (1.2 cm, $2.5^{\circ}$) and strong instruction compliance (4.8/5), indicating reliable perception-to-action grounding for unseen shapes. Under ambiguous spatial relations (e.g., “place next to”), performance remained high—94\%/90\% success, compliance 4.5, and slightly larger errors (2.1 cm, $4.2^{\circ}$) reflecting softer constraints. For absolute 6D pose execution, tighter spatial demands and calibration sensitivity reduced success to 92\%/87\%, yet errors stayed low (2.8 cm, $6.0^{\circ}$) with good compliance (4.2). The most challenging relative pose tasks yielded 91\%/85\% success, 4.4 compliance, and errors of 3.1 cm, $6.8^{\circ}$, consistent with compounded uncertainties in relational frames.

Across all tasks, dual-arm coordination was robust despite differing grippers and roles: the UR5e (Robotiq) typically retrieved source blocks while the UR10e (Barrett) stabilized and aligned placements. Overall, GF-VLA consistently interprets diverse spatial commands—including ambiguous and reference-relative instructions—and generalizes to novel shapes and unseen block combinations under real-world constraints.

\subsection{Task manipulation assessment}

\begin{table}[t]
\centering
\caption{Generalization Performance on Dual-Arm Policy Transfer Tasks}
\begin{tabular}{lccc}
\toprule
\textbf{Assembled Object} & \textbf{TSR} & \textbf{BCS} & \textbf{PTR} \\
\midrule
Etter: “VLM”             & 0.90 & 4.6 $\pm$ 0.4 & 0.85 \\
Letter:  randomized poses  & 0.87 & 4.4 $\pm$ 0.5 & 0.83 \\
Letter:  distractor present & 0.88 & 4.5 $\pm$ 0.5 & 0.84 \\
Tower: square tower                & 0.93 & 4.7 $\pm$ 0.3 & 0.89 \\
Tower: asymmetric stack            & 0.90 & 4.5 $\pm$ 0.4 & 0.87 \\
Tower: shifted viewpoint           & 0.92 & 4.6 $\pm$ 0.4 & 0.88 \\
\midrule
\textbf{Overall Average}           & \textbf{0.90} & \textbf{4.55} & \textbf{0.86} \\
\bottomrule
\end{tabular}
\label{tab:generalization_transfer}
\end{table}

Table~\ref{tab:generalization_transfer} shows that GF-VLA generalizes dual-arm policies distilled from a single human demonstration to six structurally related, novel tasks, achieving an overall task success rate (TSR) of 90\% across variations in arrangement, pose, and perception. For symbolic letter assembly—requiring complex spatial configuration and semantic structure—TSR ranged 87–90\%. The mild degradation under randomized initial poses indicates that LLM-generated plans preserved high-level intent but needed limited low-level correction; nonetheless, plan transferability (PTR) remained $\ge$83\% across all letter variants, and bimanual coordination scores (BCS) were consistently high ($\ge$4.4), evidencing smooth, temporally aligned division of labor between the Robotiq and Barrett grippers.

In tower construction—more geometric and less semantically constrained—performance improved further (TSR up to 93\%, PTR up to 89\%), with negligible sensitivity to shifted camera viewpoints, highlighting robust spatial grounding. Symmetric motion primitives facilitated direct plan reuse, and coordination remained strong (BCS $\ge$4.5) with synchronized lifts and stable dual-arm placements. Overall, reusing a single demonstration to execute six novel variants with $\ge$85\% success and high coordination underscores the effectiveness of information-theoretic behavior graphs and the LLM’s abstract reasoning for policy transfer.

\section{Conclusion}
We present GF-VLA, a graph-fused vision–language–action framework that lifts Learning from Demonstration from motion imitation to task-level policy reasoning from a single human video. Shannon-information cues select informative hand–object relations, which are encoded as temporally ordered \textit{HO}/\textit{OO} graphs and fused with a language-conditioned transformer to emit behavior trees and Cartesian commands; a learned cross-hand selector assigns bimanual roles without explicit geometry. A dual-head (LLM + action) with Chain-of-Thought and self-verification yields interpretable plans and executable actions. Experiments validate the perception–planning–execution pipeline: information-theoretic graphs are stable and task-relevant; the language-guided policy is expert-aligned; the integrated system is robust to variations in shape, pose, and instruction; and learned policies generalize from one demonstration to novel assemblies and viewpoints. On block-stacking, letter-building, and tower-reconfiguration, a dual-arm platform achieved high success and halved 6D-pose error versus imitation-learning and language-only baselines, highlighting the value of fusing symbolic interaction structure with VLA reasoning. Limitations include weaker temporal segmentation under ambiguous or highly dynamic contacts, reduced ordering accuracy for parallel/symmetric steps, and reliance on calibrated camera–arm alignment within a largely static workspace. Future work targets end-to-end trainable graph extraction, long-horizon memory, extension to deformable/articulated objects, and stronger self-correction and failure recovery via verification-driven feedback.

\bibliographystyle{ieeetr}
\bibliography{main}

\begin{thebibliography}{10}

\bibitem{cao2021six}
M.~Y. Cao, S.~Laws, and F.~R. y~Baena, ``Six-axis force/torque sensors for robotics applications: A review,'' {\em IEEE Sensors Journal}, vol.~21, no.~24, pp.~27238--27251, 2021.

\bibitem{wang2024hypermotion}
J.~Wang, R.~Dai, W.~Wang, L.~Rossini, F.~Ruscelli, and N.~Tsagarakis, ``{HYPER}motion: Learning hybrid behavior planning for autonomous loco-manipulation,'' in {\em 8th Annual Conference on Robot Learning}, 2024.

\bibitem{liu2025hybridvla}
J.~Liu, H.~Chen, P.~An, Z.~Liu, R.~Zhang, C.~Gu, X.~Li, Z.~Guo, S.~Chen, M.~Liu, {\em et~al.}, ``Hybridvla: Collaborative diffusion and autoregression in a unified vision-language-action model,'' {\em arXiv preprint arXiv:2503.10631}, 2025.

\bibitem{brohan2023rt2}
A.~Brohan, N.~Chen, D.~Fu, {\em et~al.}, ``Rt-2: Vision-language-action models transfer web knowledge to robotic control,'' {\em arXiv preprint arXiv:2307.15818}, 2023.

\bibitem{kim2024openvla}
M.~J. Kim, K.~Pertsch, S.~Karamcheti, T.~Xiao, A.~Balakrishna, S.~Nair, R.~Rafailov, E.~Foster, G.~Lam, P.~Sanketi, {\em et~al.}, ``Openvla: An open-source vision-language-action model,'' {\em arXiv preprint arXiv:2406.09246}, 2024.

\bibitem{black2024pi0}
J.~Black, A.~Gokaslan, {\em et~al.}, ``Pi0: Open-ended robotic manipulation with large language models and language-conditioned skills,'' {\em arXiv preprint arXiv:2402.00100}, 2024.

\bibitem{shridhar2022cliport}
M.~Shridhar {\em et~al.}, ``Cliport: What and where pathways for robotic manipulation,'' {\em Conference on Robot Learning (CoRL)}, 2022.

\bibitem{li2018deep}
Y.~Li, ``Deep reinforcement learning: An overview,'' {\em arXiv preprint arXiv:1701.07274}, 2018.

\bibitem{peng2020learning}
X.~B. Peng, P.~Abbeel, S.~Levine, and M.~van~de Panne, ``Learning transferable robot skills with hierarchical latent variable models,'' in {\em International Conference on Learning Representations (ICLR)}, 2020.

\bibitem{jaquier2025transfer}
N.~Jaquier, M.~C. Welle, A.~Gams, K.~Yao, B.~Fichera, A.~Billard, A.~Ude, T.~Asfour, and D.~Kragic, ``Transfer learning in robotics: An upcoming breakthrough? a review of promises and challenges,'' {\em The International Journal of Robotics Research}, vol.~44, no.~3, pp.~465--485, 2025.

\bibitem{goyal2019infobot}
A.~Goyal, R.~Liu, T.~Fotiadis, and et~al., ``Infobot: Transfer and exploration via the information bottleneck,'' in {\em International Conference on Machine Learning}, pp.~2832--2842, PMLR, 2019.

\bibitem{shannon1948mathematical}
C.~E. Shannon, ``A mathematical theory of communication,'' {\em The Bell system technical journal}, vol.~27, no.~3, pp.~379--423, 1948.

\bibitem{wei2022chain}
J.~Wei, X.~Wang, D.~Schuurmans, M.~Bosma, B.~Ichter, F.~Xia, E.~Chi, Q.~Le, and D.~Zhou, ``Chain-of-thought prompting elicits reasoning in large language models,'' 2023.

\bibitem{hu2021lora}
E.~J. Hu, Y.~Shen, P.~Wallis, Z.~Allen-Zhu, Y.~Li, S.~Wang, L.~Wang, and W.~Chen, ``Lora: Low-rank adaptation of large language models,'' {\em arXiv preprint arXiv:2106.09685}, 2021.

\bibitem{merlo2025exploiting}
E.~Merlo, M.~Lagomarsino, E.~Lamon, and A.~Ajoudani, ``Exploiting information theory for intuitive robot programming of manual activities,'' {\em IEEE Transactions on Robotics}, 2025.

\end{thebibliography}

\end{document}